\documentclass[a4paper, 10pt] {article}

\newcommand{\match}{\textrm{match}}

\newcommand{\DFA}{\emph{\textrm{DFA}}}

\newcommand{\vyi}{V^\emptyset_{y^i}}
\newcommand{\vy}[1]{V^\emptyset_{y^{#1}}}

\newcommand{\rhid}[1]{R}
\newcommand{\rvid}[1]{R}

\newcommand{\av}{a}

\newcommand{\Sigmac}{\Sigma_\mathcal C}
\newcommand{\Sigmat}{\Sigma_\mathcal C}
\newcommand{\quotes}[1]{``#1\textrm{''}}

\newcommand{\vx}[1]{V^\rho_{x_{#1}}}
\newcommand{\vxi}{{V^\rho_{x_i}}}

\renewcommand{\dh}{\textrm{$\hat\delta$}}
\newcommand{\dhi}{\dh_i}

\newcommand{\defeq}{\stackrel{def}{=}}

\usepackage{latexsym}
\usepackage[english]{babel}
\usepackage{graphics}
\usepackage{graphpap}
\usepackage{color}
\usepackage{clrscode}
\usepackage{psfrag}

\usepackage{epsf}
\usepackage{wrapfig}
\usepackage{graphicx}
\usepackage{amsmath, amssymb, amscd, amsthm}
\newtheorem{definition}{Definition}

\newtheorem{lemma}{Lemma}
\newtheorem{theorem}{Theorem}
\newtheorem{example}{Example}

\newenvironment{myenumerate}{
\begin{enumerate}
  \setlength{\itemsep}{0pt}
  \setlength{\parskip}{0pt}
  \setlength{\parsep}{0pt}}
{\end{enumerate}
}

\begin{document}
\author{Esben Rune Hansen$^1$ and Henrik Reif Andersen$^1$\\\\
$^ 1$ IT University of Copenhagen,\\ 
Rued Langgaards Vej 7, DK-2300 Copenhagen S, Denmark\\
$\{$esben,hra$\}$@itu.dk}

\title{Interactive Configuration by Regular String Constraints}
\maketitle

\paragraph{Abstract}
A product configurator which is complete, backtrack free and able to
compute the valid domains at any state of the configuration can be
constructed by building a Binary Decision Diagram (BDD). Despite the
fact that the size of the BDD is exponential in the number of
variables in the worst case, BDDs have proved to work very well in
practice. Current BDD-based techniques can only handle interactive
configuration with small finite domains. In this paper we extend the
approach to handle string variables constrained by regular
expressions. The user is allowed to change the strings by adding
letters at the end of the string. We show how to make a data structure
that can perform fast valid domain computations given some assignment
on the set of string variables.

We first show how to do this by using one large DFA. Since this
approach is too space consuming to be of practical use, we construct a
data structure that simulates the large DFA and in most practical
cases are much more space efficient. As an example a configuration
problem on $n$ string variables with only one solution in which each
string variable is assigned to a value of length of $k$ the former
structure will use $\Omega(k^n)$ space whereas the latter only need
$O(kn)$. We also show how this framework easily can be combined with
the recent BDD techniques to allow both boolean, integer and string
variables in the configuration problem.

\section{Introduction}
Interactive configuration is a special Constraint Satisfaction Problem
(CSP), where a user is assisted in configuration by interacting with a
\emph{configurator} -- a computer program. In configuration the user
repeatedly chooses an unassigned variable and assigns it a value until
all variables are assigned.  The task of the configurator is to state
the valid choices for each of the unassigned variable during the
configuration. The set of valid choices for an unassigned variable $x$
is denoted the \emph{valid domain} of $x$ \cite{onlineConfig1},
\cite{onlineConfig2}.

As an example consider the problem of assigning values to the
variables $x_1, x_2$ and $x_3$ where $x_1 \in \{1,\ldots,5\}$ and
$x_2, x_3 \in \{1,\ldots,10\}$ with the requirement that $x_1 = 1 \lor
x_1 = 2 \lor x_2 = 2$ and $x_2 = x_3$. Initially the user can choose
to assign a value from $\{1,\ldots,5\}$ to $x_1$ or assign a value
from $\{1,\ldots,10\}$ to $x_2$ or $x_3$. Suppose now the user assigns
$3$ to $x_3$. In this case the valid domain of $x_2$ is $\{3\}$ and
the valid domain of $x_1$ is $\{1,2\}$. We obtain the requirement $x_2 = 3$ 
by $x_2 = x_3$ and $x_3 = 3$. Further we obtain $x_1 \in \{1,2\}$ by 
$x_1 = 1 \lor x_1 = 2 \lor x_2 = 2$ and $x_2 = 3$.

The valid domain of each unassigned variable has to be updated every
time a value is assigned to a variable as the assignment might make
other assignments invalid as in the example above. The user interaction
with the configurator has to be real-time which in practice means that
the configurator has to update the valid domains within 250
milliseconds \cite{250millisec}. Calculating the valid domains is
NP-hard since it can be used to solve 3SAT. However by making an
off-line construction of a Binary Decision Diagram that represents the
constraints we are able to keep the computation time polynomial
in the size of the BDD. The BDD constructed can be exponentially large, but
in practice BDDs have proved themselves to be far from exponential in
size for many configuration problems.

As BDDs use binary variables to represent the domains of the
variables we normally assume small finite domains. In this paper we
will consider the case of variables that take strings as their values,
hence their domain might not be finite. Therefore the standard BDD
approach will not be able to handle the problem. 

As an example suppose that a user has to fill in a form were there is
a lot of constraints on the data. Consider a CSP with the variables
\textsf{phone}, \textsf{country}, \textsf{zip} and \textsf{district}
along with the following constraints:

\begin{itemize}
\item[I] The prefix of \textsf{phone} is ``+45'' $\iff$
\textsf{country} = ``Denmark''
\item[II] \textsf{country} = ``Denmark'' $\implies$ \textsf{zip} has
four digits
\item[III] \textsf{zip} = ``2300'' $\land$ \textsf{country} =
``Denmark'' $\iff$ \textsf{district} = ``Copenhagen S''
\end{itemize}
Suppose in the CSP above that the user entered \textsf{district} =
``Copenhagen S''.  This restricts the valid domain of \textsf{zip}
to the singleton set \{``2300''\} and the valid domain of
\textsf{country} to \{``Denmark''\} by (III). The valid domain of
\textsf{phone} is decreased to the set of strings which has
``+45'' as a prefix by (I).

Suppose instead that the user have entered \textsf{phone} = ``+45
23493844''. This decreases the valid domain of \textsf{country} to
\{``Denmark''\} by (I), and the valid domain of \textsf{zip} to
strings consisting of 4 digits. Actually this restriction will be
performed as soon as the user have entered ``+45'', since every completion
of \textsf{phone} achieved by appending additional letters at the
end of \textsf{phone} still will have ``+45'' as a prefix.

\section{Related Work}
It has recently been proposed to introduce global constraints that
require that the variables in the CSP considered in some order has to
belong to a regular language, supposing that the domain of each
variable is contained in the alphabet of the regular language
\cite{pesant}. This approach has this year (in 2006) been extended to global
constraints where the variables of the CSP have to belong to a specified
context-free grammar \cite{grammar1}\cite{grammar2}. Both results give
algorithms for ensuring generalized arc consistency which corresponds 
to valid domains in the case of interactive configuration. 

Since the value of the variables they consider is one letter in the
alphabet of the regular language, all words in the regular language
they consider have some fixed length. 

The type of constraint considered in
this paper supports variables that consist of
any number of letters. Further it allows formulas that are
multiple membership constraints connected by the boolean operators
$\land, \lor$ and $\lnot$.

\section{Preliminaries}\label{problem}
Consider a CSP stated as $\mathcal C = (\mathcal X,\Sigma, \mathcal F)$. By
$\mathcal{X} = \{x_1,x_2,\ldots,x_n\}$ we denote the variables of the
problem. By $\Sigma$ we denote an alphabet. By $\mathcal{F} =
\{f_1,\ldots,f_o\}$ we denote formulas written using the following
syntax $$f ::= f \lor f \mid \lnot f \mid \match(x,\alpha),$$ where
$\alpha$ is a regular expression over $\Sigma$. The expression
match$(x,\alpha)$ is true if and only if $x \in
L(\alpha)$, where $L(\alpha)$ is the language defined by the regular
expression $\alpha$. We use $f \land g, f \Rightarrow g$ and $f
\Leftrightarrow g$ as shortcuts for $\lnot (\lnot f \lor \lnot
g),\lnot f \lor g$ and $(f \Rightarrow g) \land (g \Rightarrow f)$
respectively.

Regular expression are written on the syntax:
$$\alpha ::= \alpha\alpha \: \big| \: \alpha | \alpha
\:\big|\: \alpha*$$ listed in increasing order of strength of
binding. The expression $\alpha*$ is zero or more repetitions of
$\alpha$. The expression $\alpha\alpha$ is the concatenation of two
regular expressions. The expression $\alpha_1|\alpha_2$ means that
either $\alpha_1$ or $\alpha_2$. For instance $L\big(a|c|(abc*)d\big)
= L\big(\big(a|c|(ab(c*))\big)d\big) = \big\{$``ad'', ``cd'', ``abd'',
``abcd'', ``abccd'', ``abcccd'', $\ldots\big\}$. We further use ``.''
as a shortcut for any letter in $\Sigma$ -- i.e. ``$w_1|w_2|\ldots
|w_{|\Sigma|}$'' where $\{w_k \mid 1 \le k \le |\Sigma|\} =
\Sigma$.

In the example where user had to fill in some data the restriction (I)
from last section would be stated as:
$$match(\textsf{phone},\textrm{``+45.*''}) \iff
match(\textsf{country},\textrm{``Denmark''})$$ where \textsf{phone}
and \textsf{country} are two variables in $\mathcal X$.

We denote by $\rho = \{(x_1,w_1),\ldots,(x_n,w_n)\}$ a complete
assignment of the values $w_1, \ldots , w_n \in \Sigma^*$ to the
variables $x_1,\ldots,x_n$ that is all the variables in $\mathcal X$.
We define $\Sigma^*$ in the usual way as
$\epsilon\cup\Sigma\cup\Sigma^3\cup\cdots$.  The set of solutions to
$\mathcal C$ is the set of assignments to $\mathcal X$ that satisfy
all formulas in $\mathcal F$, stated formally:

$$sol(\mathcal C) = \{\rho \mid \rho \models f
\textrm{ for all } f \in \mathcal F\}$$

\begin{definition}[Valid Domains]\label{validDomainsDefinition}
  The valid domain of $x_i \in \mathcal X$ relative to an assignment
  $\rho$, denoted $V^\rho_{x_i}$, is the set of values $w \in
  \Sigma^*$ for which appending $w$ to the current assignment to $x_i$
  can be extended to a solution to $\mathcal C$ by appending an
  appropriate string to values to the assignment to the remaining variables
  $\mathcal X\setminus \{x_i\}$. Stated formally:
$$\vxi =\big\{w \in \Sigma^* \mid \exists \rho': \rho'(x_i) = w \land \rho\rho' \in sol(\mathcal C)\}$$
\noindent where $\rho$ and $\rho'$ are assignments to $\mathcal X$ and
the concatenation $\rho\rho'$ is defined by $\rho\rho' =
\{(x_1,\rho(x_1)\rho'(x_1)),\ldots,(x_n,\rho(x_n)\rho'(x_n)) \}$
\end{definition}
The following theorem will be proved in the next section:
\begin{theorem}\label{validDomainIsRegular}
For any $x \in \mathcal X$ and any assignment $\rho$ to $\mathcal X$
it holds that $V^\rho_x$ is a regular language.
\end{theorem}
The goal of this paper is to construct a data structure that based on
a CSP $\mathcal C = (\mathcal X,\Sigma, \mathcal F)$ support three
operations: 
\begin{itemize}
\item[]\proc{Build}$(\mathcal C)$ that constructs the data
structure from $\mathcal C$,
\item[]\proc{Append}$(x_i,w)$ that updates $\rho$ by setting
  $\rho(x_i)$ to $\rho(x_i)w$ and makes the data structure conform to
  the new $\rho$, and
\item[]\proc{ValidDomain}$(x_i)$ that returns a regular expression
  that corresponds to the valid domain of $x_i$ on $\rho$ that is a
  regular expression $\alpha$ for which $L(\alpha) = \vxi$.
\end{itemize}
As the two latter algorithms has to be used during user interaction
the goal is to make these two operations run as fast as possible
without using too much space.

One might consider a fourth operation \proc{Complete}$(x_i)$ that
indicates that there will be no more updates to the value of some
string variable which will imply an additional reduction of the valid
domains. In the context of form validation this corresponds to the
event that the user hits the return key or leaves the current input
field. In Section \ref{completingString} we show that this extra
functionality easily can be supported by the three operations
already mentioned.

In order to check whether $w \in L(\alpha)$
we use a deterministic finite automaton (DFA). 
We denote DFAs deciding the regular expressions that
occurs in $\mathcal F$ by the name \emph{match-DFAs}.

\section{A Solution based on a single DFA}\label{bigDfaSection}
In this section we will prove that $\vxi$ is a regular language.
However we want to do more than that. We will present a construction
of a DFA that for any $x_i \in \mathcal X$ and any assignment $\rho$ to
$\mathcal X$, can be turned into a DFA deciding $\vxi$. This proves
that $\vxi$ is a regular language but the data structure that will be
presented in this section uses too much space to be of any practical
use. However it gives us a good starting point for making a smaller
efficient data structure supporting the operations \proc{Build},
\proc{Append} and \proc{ValidDomain} mentioned in the last section.

The DFA we want to construct is denoted $M_\mathcal C$, and is the DFA
deciding a language we denote $L_\mathcal C$. We will now spend some
time on defining the language $L_\mathcal C$. The basic property of
$L_\mathcal C$ is that:
\begin{equation}\label{lcEqualsSol}
w \in L_\mathcal C \iff \rho_w \in sol(\mathcal C)
\end{equation}
where $w$ is a word that induces the assignment
$\rho_w$, where the meaning of induces will be defined in \eqref{wpCorrespondence}.

Intuitively we make the alphabet of $L_\mathcal C$, denoted
$\Sigmac$, consist of all possible \proc{Append}-operations
More formally stated $\Sigmac \subseteq (\Sigma
\cup \{\epsilon\})^n$ where each letter in $\Sigmac$ only contain one element
different from $\epsilon$ that is:
\begin{equation}\label{SigmaC}
\Sigmac \defeq \bigcup_{1 \le i \le n}\bigcup_{w \in \Sigma}\big\{(\;\underbrace{\epsilon,\ldots,\epsilon}_{i-1},w,\underbrace{\epsilon,\ldots,\epsilon}_{n-i}\;)\big\}
\end{equation}
Every word $w$ in $L_\mathcal C$ is a concatenation of
letters from $\Sigmac$ that is $L_\mathcal C \subseteq \Sigma_\mathcal C^*$. We say that:
\begin{equation}\label{wpCorrespondence}
w = w_1\cdots w_k\; \textrm{ induces }\;\rho_w = \{(x_1,w_{1,1}\cdots w_{k,1}),\ldots,(x_n,w_{1,n}\cdots w_{k,n})\}
\end{equation}
where $w_{l,i}$ denotes the $i$th element in the letter $w_l \in
\Sigma_\mathcal C$ and $1 \le l \le k$ and $1 \le i \le n$ and
$\rho_w$ is an assignment to $\mathcal X$. Note that for any $w =
w_1\cdots w_k$ every word $w'$ that consist of the exactly the letters
$w_1,\cdots,w_k$ permuted in a way that maintains the ordering of
$w_{i,1}, \ldots w_{i,k}$ for every $1\le i \le n$ we have $\rho_{w'}
= \rho_w$. Hence every assignment $\rho_w$ corresponds to exactly the
$\frac{w!}{|\rho_w(x_1)|!\cdots|\rho_w(x_n)|!}$ different words. For
convenience we will in the following, when we use $w$ and $\rho_w$ in
the same calculations, implicitly assume that $w$ induces $\rho_w$ as
defined in \eqref{wpCorrespondence}.

\begin{example}
Consider the CSP where $\mathcal X = \{x_1,x_2,x_3\}$ and $\Sigma = \{a,b\}$.
In this case $$\Sigma_\mathcal C = 
\{(a,\epsilon,\epsilon),
(b,\epsilon,\epsilon),
(\epsilon,a,\epsilon),
(\epsilon,b,\epsilon),
(\epsilon,\epsilon,a),
(\epsilon,\epsilon,b)\}
$$ and for instance does the word
$w = 
(a,\epsilon,\epsilon)
(\epsilon,\epsilon,a)
(b,\epsilon,\epsilon)
(a,\epsilon,\epsilon)
$
induce the assignment $\rho_w = \{(x_1,aba),(x_2,\epsilon),(x_3,a)\}$, and 
so does for instance 
$w' = 
(a,\epsilon,\epsilon)
(b,\epsilon,\epsilon)
(a,\epsilon,\epsilon)
(\epsilon,\epsilon,a)
$
and 
$w'' = 
(a,\epsilon,\epsilon)
(b,\epsilon,\epsilon)
(\epsilon,\epsilon,a)
(a,\epsilon,\epsilon)
$.
In the case of $w$, \eqref{lcEqualsSol} becomes:
$$
(b,\epsilon,\epsilon)
(\epsilon,\epsilon,b)
(b,\epsilon,\epsilon)
(a,\epsilon,\epsilon) \in L_\mathcal C \iff
\{(x_1,aba),(x_2,\epsilon),(x_3,a)\} \in sol(\mathcal C)
$$
Note however that for instance $w''' = 
(b,\epsilon,\epsilon)
(a,\epsilon,\epsilon)
(\epsilon,\epsilon,a)
(a,\epsilon,\epsilon)
$ does not induce $\rho_w$, since $\rho_{w'''} = \{(x_1,baa),(x_2,\epsilon),(x_3,a)\}$.
\end{example}

Hence if we can make a DFA that decides $L_\mathcal C$ this DFA can be
used to decide for any assignment $\rho$ whether $\rho \in
sol(\mathcal C)$. In the following we will construct such a DFA and we
will show how we based on this construction for any $\vxi$ can make a
DFA that decides the language $\vxi$ thereby showing that $\vxi$ is a
regular language. Before we begin the construction we formally define
a DFA:

\begin{definition}[DFA]\label{DFAdef}
  A deterministic finite automaton $DFA = (Q, \Sigma, \delta, s, A)$,
  has a finite set of states $Q$, a transition function $\delta: Q
  \times \Sigma \rightarrow Q$, where $\Sigma$ is some alphabet,
  a starting state $s \in Q$ and a set of accepting states $A
  \subseteq Q$. We use $\hat\delta(s,w)$ as a shorthand for
  $\delta(\cdots\delta(\delta(q,w_1),w_2),\ldots,w_l)$, where
  $(w_1,\ldots,w_l)$ are the letters of the $w \in \Sigma^*$.  If
  $q = s$ we write $\hat\delta(q,w)$ as $\hat\delta(w)$.  
\end{definition}

\begin{definition}[Reachability in a DFA]
  In a DFA $M=(Q, \Sigma, \delta, s, A)$ a state $q$ is reachable from
  a state $p$ by the string $w \in \Sigma^*$ if and only if
  $\hat\delta(p,w) = q$.  In particular any state is reachable from
  itself by the empty string. The state $q$ is reachable from $p$ if
  and only if $q$ is reachable from $p$ by some string. We say that a
  state is reachable in $M$ if it is reachable from the source.
\end{definition}

In the rest of this paper we will use the notation $p \leadsto q$ to
denote that $q$ is reachable from $p$. We will also assume that
$M_\gamma = (Q_\gamma, \Sigma_\gamma, \delta_\gamma, s_\gamma,
A_\gamma)$ for any subscript $\gamma$.

In the rest of this section we will do the following.  First we
construct the DFA $M_\mathcal C$ based on the match-DFAs of $\mathcal
F$. We then reduce the DFA $M_\mathcal C$, by replacing the alphabet
and defining $A_\mathcal C$. We thereby obtain that $M_\mathcal C$
decides the language $L_\mathcal C$ where $(q_1,\ldots,q_m) \in
Q_\mathcal C$ and $(w_1,\ldots,w_n) \in \Sigma_\mathcal C$. Finally we
show how we can turn $M_\mathcal C$ into an automaton deciding $\vxi$
by changing the source and the alphabet in $M_\mathcal C$.

\begin{figure}[ht]
\begin{center}
\input{simpledfa1.tex}\includegraphics[width=3.5cm]{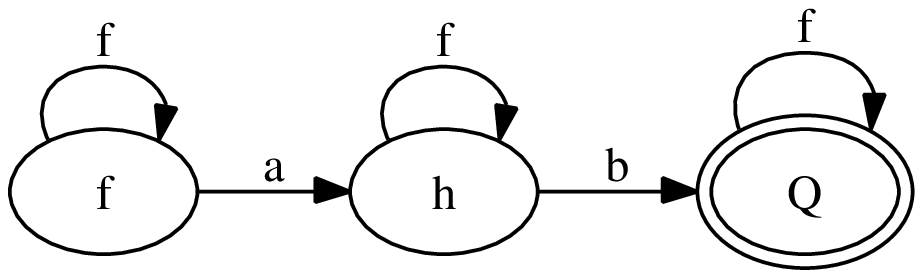}\hspace{0.3cm}
\input{simpledfa2}\includegraphics[width=3.5cm]{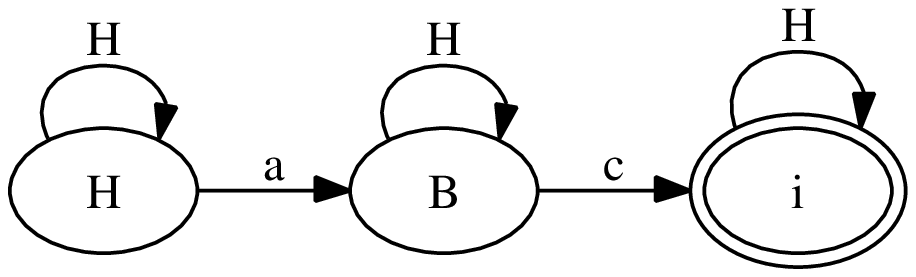}

\end{center}
\caption{DFAs on $L(\textrm{``ab''})$ and $L(\textrm{``ac''})$.}
\label{simpleSmallDfas}
\end{figure}

After this brief overview we begin the actual construction. We start
by constructing $M_\mathcal C$. This construction can be divided into
three steps:

\begin{enumerate}
\item For every match-expression $match(x_i,\alpha)$ that occurs in
  $\mathcal F$ we construct a match-DFA that decides the regular
  language $L(\alpha)$. We denote these match-DFAs $M_1,\ldots, M_m$,
  where $M_j$ is the match-DFA deciding the regular expression in the
  $j$th match-expression in $\mathcal F$, assuming some order on the
  match-expressions in $\mathcal F$. We define the mapping $I:
  \{1,\ldots,m\} \rightarrow \{1,\ldots,n\}$ such that $x_{I_j}$ is
  the variable that occurs in the $j$th match-expression.

\item For every state $q$ in the DFAs $M_1,\ldots, M_m$ we add a
  self-looping transition on the empty string $\epsilon \notin \Sigma$
  i.e. the transition $\delta(q,\epsilon) = q$. This results in DFAs
  as the ones shown in Figure \ref{simpleSmallDfas}
\item We construct a DFA $M_\mathcal C = (Q_\mathcal C, \Sigmat, \delta_\mathcal C, s_\mathcal C, A_\mathcal C)$ defined by:
\begin{itemize}
\item[$Q_\mathcal C$]$= Q_1 \times \cdots \times Q_m$ 
\item[$s_\mathcal C$]$= (s_1, \ldots ,s_m)$
\item[$\delta_\mathcal C $]$:\delta_\mathcal C((q_1,\ldots,q_m),(w_1,\ldots,w_n)) =
  (\delta_1(q_1,w_{I_1}), \ldots, \delta_m(q_n,w_{I_n}))$ where $(q_1,\ldots,q_m) \in Q_\mathcal C$ and $(w_1,\ldots,w_n) \in \Sigma_\mathcal C$
\item[$A_\mathcal C$]$ = \{(q_1,\ldots,q_m) \in Q_\mathcal C \mid \{(y_1,(q_1 \in
  A_1)),\ldots,(y_m,(q_m\in A_m))\} \models f[match(x_{I_j},\alpha_j)
  \gets y_j] \textrm{ for all }f \in \mathcal F\}$
where we by $f[match(x_{I_j},\alpha_j) \leftarrow y_j]$ mean the
formula $f$ where every match-expression on the form
$match(x_{I_j},\alpha_j)$ is replaced by the boolean variable $y_j$.

\end{itemize}
The definition of $Q_\mathcal C$ and $s_\mathcal C$ should be
straightforward. The definitions of $\delta_\mathcal C$ and $a_\mathcal C$ need
some explanation.
\end{enumerate}
In order to explain the definition of $\delta_\mathcal C$ we break it
down to four steps:
\begin{enumerate}
\item Since every state in $M_\mathcal C$ is a vector of $m$ states
a straightforward definition of $\delta_\mathcal C$ would
be on the alphabet $\Sigma^m$ on vectors on $m$ letter. Making
every transition correspond to taking exactly one step in each
of the $m$ underlying DFAs.
\item For our use we need to ensure that we follow transitions on the
  same letter in every set of DFAs that evaluates the same variable.
  This is ensured by using the mapping $I: \{1,\ldots,m\} \rightarrow
  \{1,\ldots,n\}$ defined earlier in the Section. The mapping $I$ is
  used to map every vector of letters in $(w_1,\ldots,w_n)\in\Sigma^n$
  to a vector $(w_{I_1},\ldots,w_{I_m}) \in\Sigma^m$ where $w_{I_i} =
  w_{I_j}$ if the two match-DFAs $M_i$ and $M_j$ evaluates the same variable.
\item By extending the alphabet $\Sigma^n$ to $(\Sigma \cup
  \{\epsilon\})^n$ we make it possible to make movements that
  corresponds to appending a letter to the value of a \emph{subset} of
  the variables.
\item Finally we replace the alphabet $(\Sigma \cup \{\epsilon\})^n$
  by $\Sigma_C$ as defined in \eqref{SigmaC} -- that is we remove all
  letters from the alphabet that does not correspond to appending a
  letter to the value of \emph{exactly} one variable.
\end{enumerate}
The above four steps are described in terms of the alphabet and not in
terms of transitions. However by exchanging letters above we
implicitely mean that the definitions of the transitions are
exchanged as well. If we for instance exchanged a word $w_1 \in \Sigma_1$
by $w_2 \in \Sigma_2$ the transition $\delta(p,w_1) = q$ would be exchanged
by the transition $\delta(p,w_2) = q$.
 
\begin{figure}[ht]
\begin{center}
\input{simplebigdfa.tex}
\includegraphics[width=10.0cm]{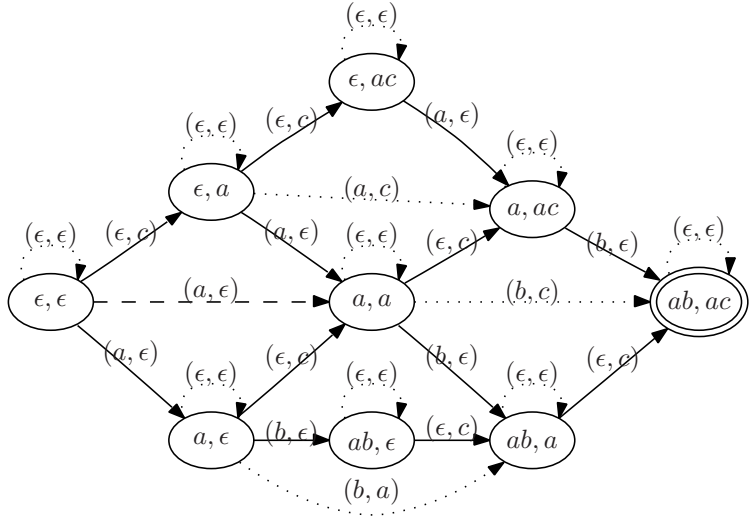}
\end{center}
\caption{The DFA $M_\mathcal C$ built based on the DFAs
  corresponding to $L($``ab''$)$ and $L($``ac''$)$ showed in Figure
  \ref{simpleSmallDfas}}\label{simpleBigDfa}
\end{figure}

\begin{example}
  Consider the example of a DFA $M_\mathcal C$ in Figure
  \ref{simpleBigDfa} based on the two match-DFAs from Figure
  \ref{simpleSmallDfas}. In figure \ref{simpleBigDfa} we have
  indicated all transitions corresponding to taking a single move in
  both of the two match-DFAs by arrows.  

  For all DFAs in this paper accepting states are indicated by double
  circles and the source is assumed to be the leftmost state in the
  graph. Further each state are labeled with the regular expression
  corresponding to the state, that is every state $q$ is labeled with
  the regular expression $\alpha$ for which $w \in L(\alpha) \iff
  \dh(w) = q$. When the alphabet is of the DFA is a subset of
  $\Sigma^2$ we label the states by two regular expression $\alpha$
  and $\beta$ such that $w_1 \in L(\alpha) \land w_2 \in L(\beta) \iff
  \dh(w_1w_2) = q$

  If the two match-DFAs are based on match-expressions on different
  variables, $\delta_\mathcal C$ is only defined for the solid
  transitions in figure \ref{simpleSmallDfas}. If the two DFAs are
  based on match-expression on the same variable $\delta_\mathcal C$
  is only defined for the dashed transition. In the latter case only
  two states are reachable from the source of the DFA,
\end{example}

In order to explain the definition of $A_\mathcal C$ we define the
transition function $\dh_\mathcal C$ as:
\begin{equation}\label{dhcDef}
\dh_\mathcal C(w) = \big(\dh_1(\rho_w(x_{I_1})),\ldots,\dh_n(\rho_w(x_{I_n}))\big)
\end{equation}
where $w$ is a word that induces $\rho_w$. Note that this
definition complies with the definition of $\dh$ in the definition of
a DFA though it differs in syntax. 

Our goal is to make $M_\mathcal C$ decide $L_\mathcal C$. In order for this
to be the case $A_\mathcal C$ has to satisfy
\begin{equation}\label{threeEquations}
\dh_\mathcal C(w) \in A_\mathcal C \iff w \in L_\mathcal C \iff \rho_w \in sol(\mathcal C)
\end{equation}
that is 
$$A_\mathcal C = \{q \in Q_\mathcal C\mid \exists w \in \Sigma^*_\mathcal C: \dh_\mathcal C(w) = q \land \rho_w \in sol(\mathcal C)\}$$
Note by \eqref{threeEquations} that for each $q \in
Q_\mathcal C$ the statement $\rho_w \in sol(\mathcal C)$ either holds
for all $w$ for which $\dh_\mathcal C(w) = q$ or for none
of these $w$s. This is due to the fact that all the $w$s for which
$\dh_\mathcal C(w) = q$ corresponds to the same set of
states in the match-DFAs: $(q_1,\ldots,q_m) = q$ and hence
will evaluate the match-expressions in $\mathcal F$ to the same boolean values.

To check for some $q \in Q_\mathcal C$ whether there exists a $w \in
\Sigma^*_\mathcal C$ for which $\dh_\mathcal C(w) = q$ is a simple
task but checking whether $\rho_w \in sol(\mathcal C)$ holds for every
$w$ for which $\dh_\mathcal C(w) = q$ requires some explanation.

Every $j$th match-expression in $\mathcal F$ evaluated by the
match-DFA $M_j$ is a term that either is $true$ or $false$ depending
on whether the current state in the $M_j$ is accepting or not. Every
state $q$ in $M_\mathcal C$ corresponds to the combination of states
$(q_1,\ldots,q_m) = q$ in the match-DFAs $M_1,\ldots,M_m$. Because of
this we might intuitively consider every match-expression as a boolean
variable. Let us denote the boolean variables $y_1,\ldots,y_m$, and
let $y_j$ correspond to the $j$th match-expression in $\mathcal F$ for
$1 \le j \le m$. Observe that every state $q = (q_1,\ldots,q_m) \in
Q_\mathcal C$ can be conceived as a complete assignment of boolean
values to such $y_1,\ldots,y_m$ by acceptance/rejection of
$q_1,\ldots,q_m$ by $M_1,\ldots,M_m$ respectively.  If this complete
assignment satisfies every formula $f \in \mathcal F$, then we have
for every $\dh(w) = q$ that $\rho_w \in sol(\mathcal C)$, otherwise we
have for every $\dh(w) = q$ that $\rho_w \notin sol(\mathcal C)$.

We restate this in formal terms. We first define the assignment
$$\tau_q = \{(y_1,(q_1 \in A_1)),\ldots,(y_m,(q_m\in A_m))\}$$
to the boolean variables $y_1,\ldots,y_m$. We let $\alpha_j$ be the
regular-expression in the $j$th match-expression and obtain
$$\rho_w \in sol(\mathcal C) \iff \tau_{\dh(w)} \models f[match(x_{I_j},\alpha_j) \leftarrow y_j] \textrm{ for all } f \in \mathcal F$$
where we by $f[match(x_{I_j},\alpha_j) \leftarrow y_j]$ we mean the
formula $f$ where every match-expression on the form
$match(x_{I_j},\alpha_j)$ is replaced by the boolean variable $y_j$.
Using equation \eqref{threeEquations} this can be rewritten as:
$$A_\mathcal C = \{q \in Q_\mathcal C \mid \tau_q \models f[match(x_{I_j},\alpha_j) \leftarrow y_j] \textrm{ for all } f \in \mathcal F\}$$
Checking for some $q$ whether $\tau_q \models
f[match(x_{I_j},\alpha_j) \leftarrow y_j]$ can be done by
simply plugging in some values in the boolean formula $f$ and checking
whether this makes $f$ true or false.

Having explained $\delta_\mathcal C$ and $A_\mathcal C$ we now
consider how to turn $M_\mathcal C$ into a DFA that decides
$\vxi$. We start by stating the definition of valid domains
(Definition \ref{validDomainsDefinition}) in term of the language
$L_\mathcal C$ as:
$$\vxi = \{w \in \Sigma^*  \mid \exists w_\mathcal C \in L_\mathcal C : \rho_{w_\mathcal C}(x_i) = \rho(x_i)w\}$$
If we want to change $M_\mathcal C$ such that it decides $\vxi$ we have to do two things:
\begin{enumerate}
\item Set the source in $M_\mathcal C$ to $\dh_\mathcal C(w)$, where $w \in \Sigma_C$ is the word corresponding to $\rho$
\item Project the alphabet on $x_i$ -- that is, replace every letter $w = (w_1,\ldots,w_n) \in
  \Sigmac$ by $w_i \in \Sigma \cup \{\epsilon\}$
\end{enumerate}
Note that the second step turn all transitions on $w$ for which $w_i =
\epsilon$ into $\epsilon$-transitions, hence we have made a
non-deterministic automaton on the alphabet $\Sigma$, deciding $\vxi$.
Using basic automata theory we obtain a corresponding DFA and the
corresponding regular expression. 

\begin{figure}[ht]
\begin{center}
$f_1:\;$\input{dfa1.tex}\includegraphics[width=3.5cm]{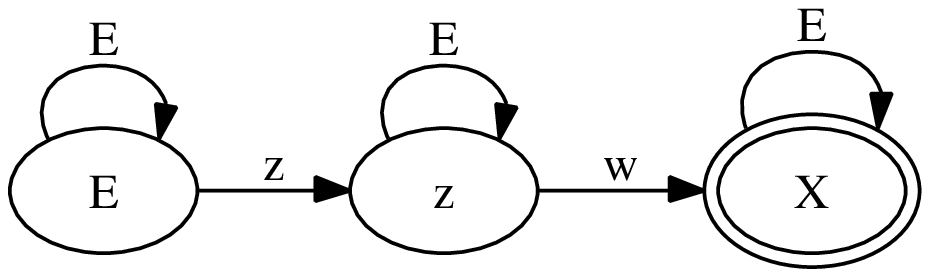}\hspace{0.3cm}$\lor$
{\input{dfa2a.tex}\includegraphics[width=5.0cm]{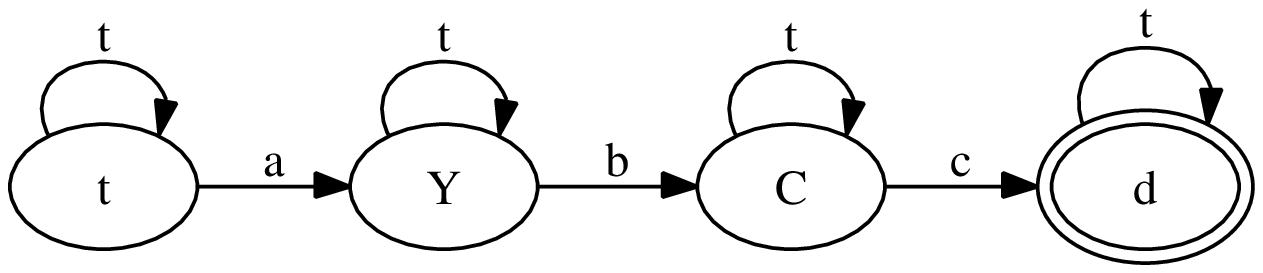}}\\
$f_2:$
{\input{dfa2b.tex}\includegraphics[width=5.0cm]{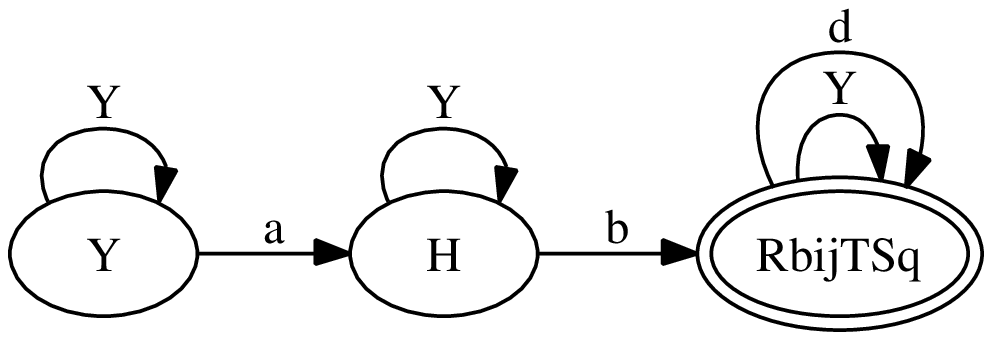}}

\end{center}
\caption{The upper two DFAs stems from $match(x_1,\textrm{``ab''})$
  and $match(x_2,\textrm{``abc''})$ respectively. The lower DFA stems
  from $match(x_2,\textrm{``abd$*$''})$}
\label{smallDfas}
\end{figure}

\begin{figure}[ht]
\begin{center}
\input{bigdfa.tex}
\includegraphics[width=10.0cm]{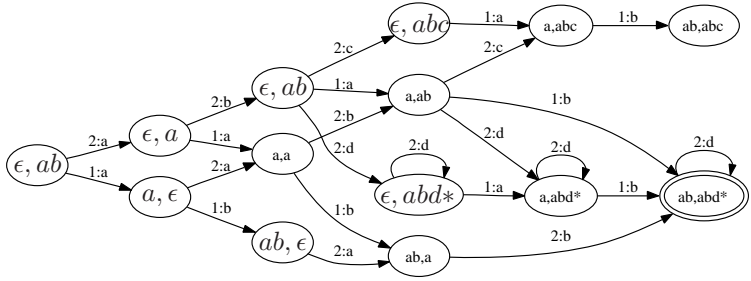}
\end{center}
\caption{A DFA $M_\mathcal C$ built on the formula $f_1 =
  match(x_2,$``abc''$) \lor match(x_1,$``ab''$) \land
  match(x_2,$``abd$*$''$)$. The transition-labels 1:$w$ and 2:$w$
  where $w \in \Sigma$ corresponds to the assignments $\rho =
  \{(x_1,w)(x_2,\epsilon)\}$ and $\{(x_1,\epsilon),(x_2,w)\}$
  respectively.  For simplicity the states corresponding to rejection
  of any of the match-expressions are not included. A DFA with all
  states are shown in Figure \ref{veryBigDfa} in the Appendix.}
\label{bigDfa}
\end{figure}

\begin{figure}[ht]
\input{vxibigdfa.tex}\includegraphics[width=12.0cm]{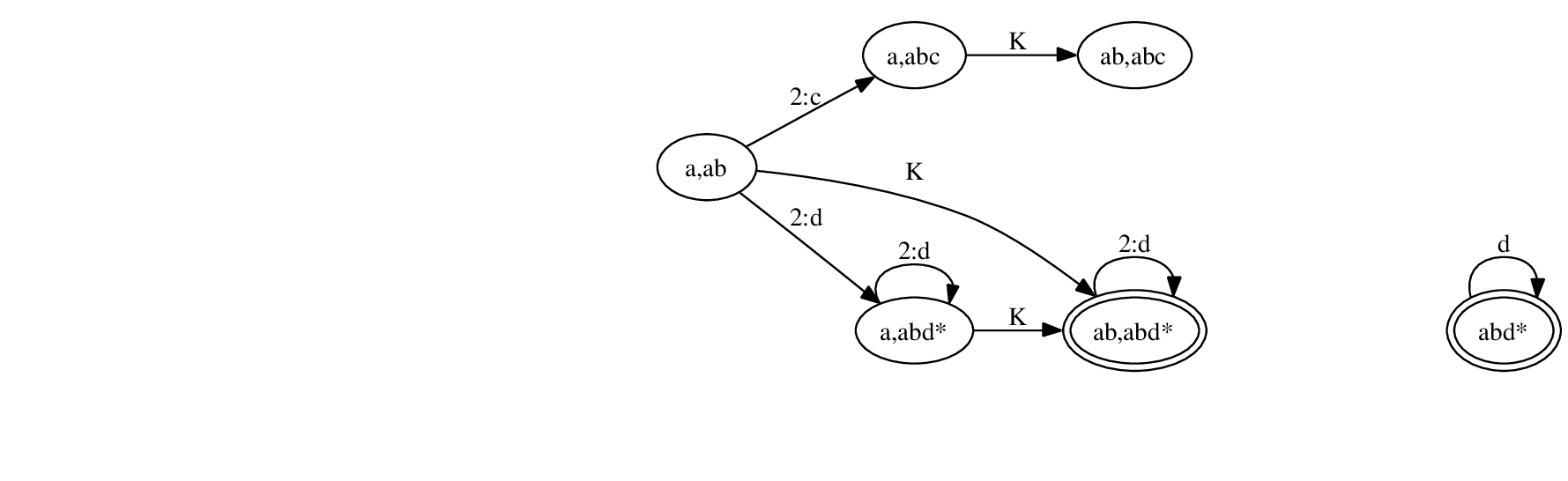}
\caption{\textbf{To the left:} The non-deterministic automaton
  deciding valid domains $\vx 2$ where $\rho =
  \{(x_1,$``a''$),(x_2,$``ab''$)\}$ derived from the DFA $M_\mathcal
  C$ in Figure \ref{bigDfa}. \textbf{To the right:} the corresponding
  DFA.\label{bigVxiDfa}}
\end{figure}

\begin{example}\label{bigDfaExample}
  Consider the example $\mathcal C = (\mathcal X,\Sigma,\mathcal F)$
  where $\mathcal X = \{x_1,x_2\}, \mathcal F = f_1, f_2$ where $f_1 =
  match(x_1,$``ab''$)\, \lor \, match(x_2,$``abc''$)$ and $f_2 =
  match(x_2,$``abd$*$''$)$. We construct the match-DFAs $M_1, M_2$ and
  $M_3$ on the regular languages L(``ab''), L(``abc'') and
  L(``abd$*$'') respectively.  To each state in $M_1, M_2$ and $M_3$
  we add $\epsilon$-transitions that are self-loops.  The resulting
  DFAs are shown in Figure \ref{smallDfas}.

  We now begin the construction of the DFA $M_\mathcal C$.  Since
  $Q_\mathcal C = Q_1 \times Q_2 \times Q_3$ this DFA will have $|Q_1|
  \cdot |Q_2| \cdot |Q_3| = 3\cdot 4\cdot 3 = 36$ states.  However not all the states
  are reachable by $s_\mathcal C$ since $\delta_\mathcal C$ is only
  defined on $\Sigma_\mathcal C = \bigcup_{w \in \Sigma}
  \big(\{(w,\epsilon)\} \cup \{(\epsilon,w)\}\big)$. The
  remaining 14 states and the transitions in $M_\mathcal C$ are shown
  in Figure \ref{bigDfa}.

  We can check for each state in $M_\mathcal C$ whether it is
  accepting by checking if its corresponding states in the match DFAs
  $M_1,\ldots,M_m$ yield an assignment to the match-expressions by
  acceptance/rejection that make $\mathcal F$ true. In this example
  only the state labeled ``$(ab,abd*)$'' is accepting. 

  Suppose now we want to calculate $\vx 2$ where $\rho =
  \{(x_1,$``$a$''$)(x_2,$``$ab$''$)\}$. We first set $s_\mathcal C =
  \dh_\mathcal C(($``$a",\epsilon)(\epsilon,$``$a")(\epsilon,$``$b"))$
  and then replace every letter $w = (w_1,w_2) \in \Sigma_\mathcal C$
  by $w_2 \in \Sigma \cup \{\epsilon\}$ that is every $(\epsilon,w_2)$
  by $w_2$ and every $(w_1,\epsilon)$ by $\epsilon$. The resulting
  non-deterministic automaton and its corresponding DFA is shown in
  Figure \ref{bigVxiDfa}.  In this example we get
  $V^{\{(x_1,\textrm{``}a\textrm{''}),(x_2,\textrm{``}ab\textrm{''})\}}_{x_2}
  = $``$d*$''.
\end{example}

\paragraph{The size of the valid domains DFA}
Though both updating $x_i$ and computing valid domains will be fast
using this solution, the size of the DFA is too large for the solution
to be of any use for larger problems. As an example a problem on $n$
variables containing a single solution
$\{(x_1,w_1),\ldots,(x_n,w_n)\}$ where $|w_i| = k$ for all $1 \le i
\le n$ the $M_\mathcal C$ will contain $\Omega(k^n)$ states. The
construction that we will achieve at the end of this paper will
contain $O(kn)$ states.

\subsection{Simulating the valid domains DFA }
In order to make a less space consuming construction we separate the
valid domains DFA into smaller DFAs, that is instead of joining all
the match-DFAs into the DFA $M_\mathcal C$ we only join match-DFAs on
the same variable.  The drawback of this approach is that we cannot
encode the boolean logic of $\mathcal F$ into the DFAs on the
variables as each of these DFAs only constitutes a partial solution to
$\mathcal F$. We therefore build a BDD on the boolean logic of
$\mathcal F$.  In this BDD every match-expression is considered as a
boolean variable. Given any combination of states in the DFAs on the
values the we can compute the value of $A_\mathcal C$ on the fly by
restricting the BDD to the acceptance and rejections of the various
values.  In this way we are able the simulate the DFA $M_\mathcal C$
by a much smaller data-structure. This structure will perform well in
terms of updating values and deciding $L_\mathcal C$ and reporting
$\vxi$. Performing this construction is the main of this paper.

In Section \ref{DFAandMDFA} we describe how to encode a set of the
DFAs on the same variable into a \emph{Multi-DFA} that can simulate
many DFAs simultaneously on the same string.  In Section
\ref{ReachabilityConstraint} we encode into every state $q$ in the
Multi-DFA which combinations of acceptance/rejection by the simulated
match-DFAs that can be reached by following transitions corresponding
to some word from $q$ in the Multi-DFA. In Section \ref{booleanLogic}
we construct the BDD taking care of the boolean logic in $\mathcal F$
as the constraint problem $\mathcal D$, where every match-expression
is replaced by a boolean variable. In Section \ref{algorithms} we
present the algorithms \proc{Build}$(\mathcal C)$,
\proc{Append}$(x_i,w)$ and \proc{ValidDomain}$(x_i)$. Finally in
Section \ref{Extensions} we consider various extensions to the data
structure.

\section{DFAs and Multi-DFAs}\label{DFAandMDFA}

By the construction of the DFA $M_\mathcal C$ in the previous section we have
ensured two properties:
\begin{enumerate}
\item  All small DFAs on the same variable are synchronized
\item  All states that cannot be a valid solution are removed
\end{enumerate}
In order to reduce the space consumption of the DFAs we will present
solution that only join match-DFAs on the same
variable. By doing this we ensure (1). In the last section we could
ensure (2) simply by minimizing the DFA. We do not have this option if
we separate DFAs on different variables since the DFAs will not have
the logic of $\mathcal F$ encoded in their structure. This problem
will be addressed in Section \ref{booleanLogic}.

Since DFAs on a single variable often will be the combination of more
than one match-DFA and since the value of one variable is not enough
to determine whether or not $\mathcal F$ is satisfied, we cannot use
acceptance and rejection in the same way as in Section
\ref{booleanLogic}.  We therefore replace the notion of accepting
states by an bit-vector denoted \emph{acceptance value} assigned to
each state containing true or false for each of the match-DFAs
accepting or rejecting for each in the current state. This is the idea
behind the following generalization of the definition of a DFA.

\begin{definition}\label{multiDFA}
  A multi-DFA (MDFA) $(Q, \Sigma, \delta, s, a)$ of acceptance size $k$, has a
  finite set of states $Q$, a transition function $\delta: Q \times
  \Sigma \rightarrow Q$, where $\Sigma$ is some alphabet, a
  starting state $s$ and an acceptance value $\av(q) \in \mathbb{B}^k$
  for every $q \in Q$. The acceptance value of a word $w$ is defined
  as $\av(\hat\delta(w)) \in \mathbb{B}^k$.
\end{definition}

Note that the definition above assigns exactly one acceptance value to
every finite string in $\Sigma^*$. Note further that an MDFA with
acceptance size $1$ is a standard DFA with the set of accepting states
$\{q \mid a(q) = (true)\}$.

As we in the rest of this paper only use the alphabet $\Sigma$
given by the CSP $\mathcal C = (\mathcal X, \Sigma, \mathcal F)$ we
will from now on not state the alphabet $\Sigma$ explicitly in our definitions of
DFAs and MDFAs. In other words we use $(Q,
\delta, s, A)$ and $(Q, \delta, s, a)$ as a shortcuts for $(Q, \Sigma,
\delta, s, A)$ and $(Q, \Sigma, \delta, s, a)$ for DFAs and MDFAs
respectively, where $\Sigma$ is the alphabet given by $\mathcal C$.

\paragraph{Construction of an MDFA} We might build an MDFA by slightly
modifying the construction of the DFA $M_\mathcal C$.  However this
might make the intermediate structure very large. Instead we
use a simple approach making a simultaneous DFS in the DFAs that has
to be joined as described in the next two bits of pseudocode.  We let
$\mu, Q, \delta, s, a, k$ and $Q_i, \delta_i, s_{i}, \av_i$, for $1
\le i \le k$ be globals.

\begin{codebox}
\Procname{$\proc{RecConstructMDFA}(q_1, \ldots, q_k)$}
\li \If $\mu(q_1, \ldots ,q_k) \textrm{ is defined }$
\li \Then
        \Return $\mu(q_1,\ldots ,q_k)$
    \End
\li create a new state $q \notin Q$
\li $Q \gets Q \cup \{q\}$
\li $\mu(q_1, \ldots, q_k) \gets q$
\li $\av(q) \gets \big((q_1 \in A_1), \ldots , (q_k \in A_k)\big)$
\li \For each $w \in \Sigma$
\li \Do $\delta(q,w) \gets \proc{RecConstructMDFA}(\delta_1(q_1,w),\ldots,\delta_k(q_k,w))$
    \End
\li \Return $q$
\end{codebox}

\begin{codebox}
\Procname{$\proc{ConstructMDFA}(\DFA_1,\ldots, \DFA_k)$}
\li $Q \gets \delta \gets a \gets \mu \gets \emptyset$
\li $s \gets \proc{RecConstructMDFA}(s_1,\ldots,s_k)$
\li \Return $(Q,\delta,s,a)$
\end{codebox}

The function $\mu$ is used to ensure a new state in the MDFA
corresponding to a position $(q_1, \ldots , q_k)$ in the DFAs is
created only once. We only create new states (by proceeding to Line 3)
if $\mu(q_1, \ldots , q_k)$ is undefined, which is the case if and
only if $(q_1, \ldots , q_k)$ has not been visited before. Otherwise
we return the previously created state that is assigned to $\mu(q_1,
\ldots , q_k)$ to the caller in Line 2. In Line 6 we by ``$q_j
\in A_j$'' mean $true$ if $q_j \in A_j$ and $false$ otherwise.

For instance the requirements $match(x_1,\quotes{abc})$ and
$match(x_1,\quotes{abd*})$ on $x_1$ will result in the
MDFA
drawn in figure
\ref{DFAfig}.

\begin{figure}[h]
\setlength{\unitlength}{0.75mm}
\begin{picture}(60,40)
\put(30,20){\circle{6}}
\put(28.7,18.3){$1$}

\put(33,20){\vector(1,0){10}}
\put(36,21){$a$}

\put(46,20){\circle{6}}
\put(44.7,18.3){$2$}

\put(49,20){\vector(1,0){10}}
\put(52,21){$b$}
\put(62,20){\circle{6}}
\put(60.7,18.3){$3$}

\put(64,22.2){\vector(1,1){7}}
\put(64,25){$c$}
\put(73,31.5){\circle{6}}
\put(71.4,29.8){$4$}

\put(64,17.8){\vector(1,-1){7}}
\put(64.4,10.7){$d$}
\put(73,8.5){\circle{6}}
\put(71.7,6.8){$5$}
\put(76,12){\oval(7,7)[r]}
\put(76,12){\oval(7,7)[t]}
\put(72.5,11.2){\vector(0,-1){0}}
\put(80.2,11){$d$}

\put(97,35){\textbf{Acceptance values}}
\put(100,28){$1: (false,false)$}
\put(100,21){$2: (false,false)$}
\put(100,14){$3: (false,true)$}
\put(100,7){$4: (true,false)$}
\put(100,0){$5: (false,true)$}

\end{picture}
\caption{The MDFA of the regular expressions: ``abc'' and ``abd*''}\label{DFAfig}
\end{figure}

Note that the state $(true,true)$ corresponding to $match(x_2,
$``$abc$''$) \land match(x_2,$``$abd*$''$)$ is not contained in the MDFA
due to the fact that $L(\quotes{abc}) \cap L(\quotes{abd*}) = \emptyset$.

Note also that this construction could be easily adapted to construct
$M_\mathcal C$ if use the alphabet $\Sigma_\mathcal C$ and following the
transition in the DFAs 

We want to make sure that the construction of the MDFA is minimal in
the number of states it is contained. In order to prove this we need
to define what means to have a minimal number of states. This can be
done by a natural generalizing the definition of a \emph{minimized}
DFA to a \emph{minimized} MDFA

\begin{definition}
A MDFA is minimized if all states in the MDFA are reachable
from $s$ and no pair of states in the MDFA are equivalent. For any pair
of nodes $p,q \in Q: p$ and $q$ are equivalent by definition if and
only if for all words $w \in \Sigma^*: \av(\hat\delta(p,w)) =
\av(\hat\delta(q,w)))$.
\end{definition}

\begin{lemma}
If the DFAs given as input to \textsc{ConstructMDFA} are minimized
then the constructed MDFA will be minimized.
\end{lemma}

\begin{proof}
 We first note that all states in $Q$ are reachable.  This is due to
the fact that every state created except $s$ will be a result of a
recursive call made at line 7. Hence every created state in the
MDFA will be assigned to a $\delta(q,w)$ for state $q$ reachable
by $s$ and some $w \in \Sigma$.

We then prove that no pair of states in the constructed MDFA is
equivalent if every $DFA_1, \ldots ,DFA_k$ is minimal.  Consider any
pair of distinct nodes $p,q \in Q$. Suppose $\mu(p_1,\ldots,p_k) =
p$ and $\mu(q_1,\ldots,q_k) = q$. Since $p \ne q$ we know by the
initial check on line 1-2 that $(p_1,\ldots,p_k) \ne (q_1,\ldots,q_k)$.
Hence for some $1\le i \le k$ we have $p_i,q_i \in Q_i$ for which $p_i
\ne q_i$. Since $DFA_i$ is minimized we know that $p_i$ is not
equivalent to $q_i$ which implies that there exists an $w \in
\Sigma^*$ for which $\av(\hat{\delta_i}(p_i,w)) \ne
\av(\hat\delta_i(q_i,w))$. This implies that
$\av(\hat{\delta}(\mu(p_1,\ldots,p_k),w)) \ne
\av(\hat\delta(\mu(q_1,\ldots,q_k),w))$ which by is the same as
$\av(\hat{\delta}(p,w)) \ne \av(\hat\delta(q,w))$. Hence $p$ and $q$ are
not equivalent.
\end{proof}

\section{Reachable acceptance values}\label{ReachabilityConstraint}

As we noticed earlier then main problem we face by not joining all
match-expression into one big DFA is that we lack the logic. We will
present a notion we call Reachable acceptance values.  The reachable
acceptance values of a state $p$ in an MDFA is the set containing exactly
the acceptance values of every state $q$ that can be reached from the
state $p$ by following zero or more transitions from $p$. Formally:

\begin{equation}\label{ReachableAcceptanceValues}
R(p) = \{\av(q) \mid p \leadsto q\}, \textrm{ where }p,q \in Q
\end{equation}
\begin{example}
\noindent The states in the MDFA on Figure \ref{DFAfig} has that following reachable acceptance values:\\
$
\begin{array}{ll}
R(1) = R(2) = R(3) &= \{(true,false),(false,true), (false,false)\},\\
R(4) &= \{(true,false),(false,false)\} \textrm{ and }\\
R(5) &= \{(false,true), (false,false)\}\\
\end{array}
$
\end{example}

The goal in this section is to compute and store the set of reachable
acceptance values for each of the states in an MDFA. When this set is
stored we can at any state of the MDFA know in advance which
acceptance values that we might end up in. Hence we can use
this to constrain the logical structure, by only allowing 
values that can be reached from the current state. The exact meaning 
of ``constraining the logical structure'' will be clear in Section \ref{booleanLogic}.

Having defined the set of reachable acceptance values we now consider how to
compute the set for every state in an MDFA in an efficient way.

\subsection{Computing the reachable acceptance values}
We start by pointing out two obvious facts about the reachable
acceptance values $R$ for the nodes in an MDFA

\begin{description}
\item[Fact 1:] If a state $p$ has transitions to exactly the states
$\{q_1,\ldots , q_l\}$ then $R(p) = \av(p) \cup R(q_1) \cup \ldots \cup
R(q_l)$
\item[Fact 2:] If two states $p,q$ belongs to the same strongly connected
component we have $R(p) = R(q)$.
\end{description}

A strongly connected component in an MDFA $(Q,\delta,s,a)$ is defined
as a set of states $C \subseteq Q$ for which it for any $p \in C$
holds that $p \leadsto q$ and $q \leadsto p$ if and only if $q \in
C$. Calculating the strongly connected components in an MDFA easily be done in
linear time in the size of the MDFA \cite{clrs}.

\begin{codebox}
\Procname{\proc{ComputeReachableAcceptanceStates}$(M)$}
\li Let $C'$ be the set of strongly connected components in $Q$
\li \For each $C_1, C_2 \in C'$
\li \Do \If $\delta(p,w) = q$ for some $p \in C_1, q \in C_2, w \in \Sigma$
\li \Then $\Gamma(C_1) \gets \Gamma(C_1) \cup C_2$
\End
\End
\li \For each $C \in C'$
\li \Do $R'(C) \gets \bigcup_{q \in C} \{\av(q)\}$\RComment{Ensure Fact 1}
\End
\li \For each $C_1 \in C'$ in reverse topological order
\li \Do $R'(C_1) = R'(C_1) \cup \bigcup_{C_2 \in \Gamma(C)} R'(C_2)$\RComment{Ensure Fact 2}
\End
\li \For each $C \in C'$
\li \Do \For each $q \in C$
\li \Do $R(q) \gets R'(C)$
\End
\End
\li \Return $R$
\end{codebox}

We assume that $M = (Q,\delta,s,a)$ is an MDFA and that initially $R =
R' = C' = \Gamma = \emptyset$.  In Line 2-4 we construct the neighbor
function $\Gamma(C)$ mapping any strongly connected component into the
set of ``children'' of the strongly connected component.  In Line 5-6
every $R'(C)$ is assigned to the set of acceptance values of the
states contained in $C$. In Line 7-8 for every connected component
$C_1$, the set $R(C_1)$ is assigned to the union of all $R'(C_2)$s for which
$C_1 \leadsto C_2$ in $C'$. Note that the topological order in $C'$ is
well defined since $C'$ is a DAG \cite{clrs}. Finally in Line 9-11 the
reachable acceptance states of the strongly connected components are
assigned to the reachable acceptance states of the states in $Q$

\section{The boolean logic of $\mathcal F$}\label{booleanLogic}
We now return to the problem of representing the boolean logic of
$\mathcal F$. In Section \ref{bigDfaSection} the boolean logic 
was contained in the DFA, in the way that every $\mathcal C$ 
in the DFA $M_\mathcal C$
constructed in Section \ref{bigDfaSection} was encoded by whether a
state was an accepting state or not.

Since we have divided the match-expressions in $\mathcal F$ into MDFAs
on each of the variables $x \in \mathcal X$ no MDFA is can in it self
decide whether $\mathcal F$ is satisfied or not. This is why the MDFAs
are neither accepting or rejecting. However if we pick a state from
each of the MDFAs this set of states is a complete assignment to the
variables in $\mathcal X$. Such a set is an accepting state if and
only if evaluating the match-expression by the rejection/acceptance of
the match-DFAs used during the construction of the MDFAs, on the
states corresponding to the states in the MDFAs, makes $\mathcal F$
true -- exactly as in Section \ref{bigDfaSection}. We denote such a
set an \emph{accepting set}. Furthermore every state in an MDFA is
\emph{valid} if it occurs in some accepting set. If it occurs in no
accepting set it is \emph{invalid}. We observe that every accepting
set of states correspond to an accepting state in $M_\mathcal C$.

In order to represent the boolean logic in $\mathcal F$ we define a
CSP $\mathcal D = (\mathcal Y, \mathbb B, \mathcal G)$ based on $\mathcal C$. The
construction of the problem has many similarities with the calculation
of the set of accepting states in $M_\mathcal C$ in Section
\ref{bigDfaSection}. The variables in $\mathcal Y$ are the same as the
$y$-variables in Section \ref{bigDfaSection} and all the constraints
$\{f[match(x_{I_j},\alpha_j) \leftarrow y_j] \mid f \in \mathcal F\}$
are constraints in $\mathcal G$. However we need some extra
constraints in $\mathcal G$ and another way to index the $y$-variables
in $\mathcal Y$ in this section, but basically this section is just an
extension of the techniques used in Section \ref{bigDfaSection}. We
will now present the notation that will be used in this section, that
will help us describe the implementation of the three operations
\proc{Build}, \proc{Append} and \proc{ValidDomain} in the next
section.

Let $\mathcal D = (\mathcal Y, \mathbb B, \mathcal G)$ be a CSP, where $\mathcal
Y = \{y_1,\ldots,y_m\}$ is a set of boolean variables and $\mathcal G$
is a set of boolean constraints on the values that can be assigned
$\mathcal Y$.  Let $\phi = \{(y_1,b_1),\ldots,(y_m,b_m)\}$, where
$y_1,\ldots,y_m \in \mathcal Y$ and $b_1,\ldots,b_m \in \mathbb B$
denote a complete assignment of the variables in $\mathcal Y$ to
boolean values, or in short: an assignment to $\mathcal Y$. We define
the solution to $\mathcal D$ by:
$$sol(\mathcal D) = \{\phi \mid \phi \models \mathcal G\}$$
where $\phi$ is an assignment to $\mathcal Y$. Further we let the formulas
$\{f[match(x_{I_j},\alpha_j) \leftarrow y_j] \mid f \in \mathcal F\}$
be a part of $\mathcal G$.

For the use of this section we will define $y^i_j$ as the $y$-variable in $\mathcal Y$
replacing the $j$th of the match-expressions on the variable $x_i$, for $1 \le i \le n$
and $1 \le j \le k_i$ where $k_i$ is the number of match-expressions
on $x_i$ that occurs in $\mathcal F$. Using this notation we can restate
$\mathcal Y$ as
\begin{equation}\label{yEquation}
\mathcal Y = \{y^1_1,\ldots,y^1_{k_1},y^2_1, \ldots, y^2_{k_2}\, \ldots\ldots ,y^n_1,\ldots,y^n_{k_n}\}
\end{equation}

Using the shortcuts $y^i = (y^i_1,\ldots,y^i_{k_i})$ and 
$b^i = (b^i_1,\ldots,b^i_{k_i})$ for every $1 \le i \le n$ where $b^i_1,\ldots,b^i_{k_i} \in \mathbb B$
we define:
$$\phi(y^i) = (y^i_1,\ldots,y^i_{k_i})$$
and 
$$y^i = b^i ::= \bigwedge_{1 \le j \le k_i} y^i_j = b^i_j$$
We further define the shortcut:
$$y^i \in B^i \stackrel{\textrm{def}}{\iff} \bigvee_{b^i \in B^i} y^i = b^i$$
where $B^i \in \mathbb B^{k_i}$.  We further denote the $j$th element
in the acceptance value of a state $q_i$ in the MDFA on $x_i$ by
$a^i_j(q_i)$ and the entire acceptance value of $q_i$ as $a^i(q_i) =
(a^i_1(q_i),\ldots,a^i_{k_i}(q_i))$, and define $R^i_j(p_i)$ as
$\{a^i_j(q_i) \mid p_i \leadsto q_i\}$ and $R^i(q_i)$ as $\{a^i(q_i)
\mid p_i \leadsto q_i\}$.

Every assignment $\rho$ to $\mathcal X$ corresponds to the assignment $\phi$ to
$\mathcal Y$ where the truth-value of $y^i_j$ corresponds to the truth value of the
$j$th match-expression of $x_i$ if where evaluating $\rho(x_i)$. More formally we say that 
$$\rho \textrm{ induces } \phi_\rho = \{(y^1,a^1(\dh_1(\rho(x_1)))),\ldots,(y^n,a^n(\dh_n(\rho(x_n))))\}$$
We want to ensure that 
\begin{equation}\label{cdEquation}
\rho \in sol(\mathcal C) \iff \phi_\rho \in sol(\mathcal D)
\end{equation}
where $\rho$ is the assignment that induces $\phi_\rho$.

The rightward implication of \eqref{cdEquation} can be satisfied by ensuring
$$\{f[match(x_i,\alpha^i_j) \leftarrow y^i_j] \mid f \in \mathcal F\}$$
by including it in $\mathcal G$, which is quite similar to what we did
in Section \ref{bigDfaSection}. 

The leftward implication in \eqref{cdEquation} was ensured in Section
\ref{bigDfaSection} by the definition of $A_\mathcal C$ and the fact
that only the accepting states that were reachable from the source of
$M_\mathcal C$ were the states $q \in A_\mathcal C$ where $q = (q_1,\ldots,q_m) =
\big(\dh'_1(w_{I_j}),\ldots,\dh'_m(w_{I_m})\big)$ for some
$w_1,\ldots,w_n \in \Sigma^*$ where $\dh'$ denotes transitions in the
match-DFAs. In this section we need to ensure the leftward implication by adding the
constraint:
\begin{equation}\label{yInR}
y^i \in R^i(s_i) \textrm{ for all } 1 \le i \le n
\end{equation}
to $\mathcal G$.
From this we get that if
$$\mathcal G = \{f[match(x_i,\alpha^i_j) \leftarrow y^i_j] \mid f \in \mathcal F\} \cup \bigwedge_{1 \le i \le n} y^i \in R^i(s_i)$$
then \eqref{cdEquation} holds.

We define the valid domains of $y^i$ by
$$\vyi = \{b^i \in \mathbb B^{k_i}\mid \exists \phi \in sol(\mathcal D): \phi(y^i) = b^i\}$$
Note that this definition is different from the definition of $\vxi$, but is
quite similar to the standard definition of valid domains as e.g. in
\cite{ValidDomainCalculation}. This version however, is specialized
for valid domains on the empty assignment and is a projection
of the valid solution onto a \emph{vector} of variables from $\mathcal Y$.

Recall the shortcut $\rho\rho'$ defined by $\rho\rho' = \{(x_1,\rho(x_1)\rho'(x_1)),\ldots,(x_n,\rho(x_n)\rho'(x_n)) \}$
used in the definition of $\vxi$ in Definition \ref{validDomainsDefinition}. Using this shortcut and that  
$$\rho \in sol(\mathcal C) \iff \phi_\rho \in sol(\mathcal D) \textrm{ where }
\phi_\rho = \{(y^1,a^1(\dh_1(\rho(x_1)))),\ldots,(y^n,a^n(\dh_n(\rho(x_n))))\}$$
we get:\\

$
\begin{array}{ll}
\vxi &= \{w \in \Sigma^*\mid \exists\rho': \rho\rho' \in sol(\mathcal C) \land \rho'(x_i)=w\}\\
     &= \{w \in \Sigma^* \mid \exists\rho': \rho\rho' \in sol(\mathcal C) \land \rho\rho'(x_i)=\rho(x_i)w\}\\
     &= \{w \in \Sigma^* \mid \exists\rho': \phi_{\rho\rho'} \in sol(\mathcal D) \land \rho\rho'(x_i)=\rho(x_i)w\}\\
     &= \{w \in \Sigma^* \mid \exists\rho': \phi_{\rho\rho'} \in sol(\mathcal D) \land \phi_{\rho\rho'}(y^i)=a^i(\dh_i(\rho(x_i)w))\}\\
     &= \{w \in \Sigma^* \mid \exists\phi \in sol(\mathcal D) \land \phi(y^i)=a^i(\dh_i(\rho(x_i)w))\}\\
     &= \{w \in \Sigma^* \mid \exists\phi \in sol(\mathcal D) \land \phi(y^i)=b^i \land b^i = a^i(\dh_i(\rho(x_i)w))\}\\
     &= \{w \in \Sigma^* \mid a^i(\dh_i(\rho(x_i)w)) \in \vyi\}\\
\end{array}
$
\\
By this we know that when $\rho \in sol(\mathcal C) \iff \phi_\rho \in sol(\mathcal D)$ is ensured
we can compute $\vxi$ using only the MDFA $M_i$ and $\vyi$. 
To obtain a DFA $(Q,\Sigma,\delta,s,A)$ deciding $\vxi$ based
on the MDFA $M_i = (Q_i,\delta^i,s_i,a^i)$ on the variable $x_i$ we
can do the following:
\begin{itemize}
\item set $Q = Q_i$ and $\delta = \delta_i$
\item set $A = \{q_i \in Q_i \mid a^i(q_i) \in \vyi\}$ 
\item set $s = \dhi(\rho(x_i))$
\end{itemize}
Note that this is very close to what was done in Section
\ref{bigDfaSection}. The main difference is that instead 
of making states accepting/rejecting at the preprocessing
we construct $A$ during the valid domain computation by
using $\vxi = \{w \mid a^i(\dh_i(\rho(x_i))w) \in \vyi)\}$.
Further we have no need to change the alphabet which is needed
in Section \ref{bigDfaSection}.

\begin{example}\label{EnsureInvariantsExample}
  Consider the CSP: $\mathcal C = (\mathcal X, \Sigma, \mathcal F)$,
  where $\mathcal X =\{x_1, x_2\}, \mathcal F = \{f_1, f_2\}, f_1 =
  match_1(x_2, $``$abc$''$)\; \lor \; match_2(x_1, \quotes a), f_2 =
  match_3(x_2, \quotes{abd*})$ and $x_1 = x_2 = \epsilon$ (Assume
  that match-expressions are ordered in increasing order of their
  subscript). We define the CSP $\mathcal D = (\mathcal Y,\mathbb B,
  \mathcal G)$. In $\mathcal D$ we have $\mathcal Y = \{y^2_1, y^1_1,
  y^2_2\}$, and disregarding the requirement \eqref{yInR} we have $\mathcal G = \{g_1, g_2\}$ where $g_1 = y^{2}_{1} \lor
  y^{1}_{1}$ and $g_2 = y^{2}_{2}$. 
We have the following facts:

$\noindent\begin{array}{llll}
\\
sol(\mathcal D)&=&\big\{&\{(y^{1}_{1},false),(y^{2}_{1},true),(y^{2}_{2},true)\},\\
&&&\{(y^{1}_{1},true),(y^{2}_{1},false),(y^{2}_{2},true)\},\\
&&&\{(y^{1}_{1},true),\;(y^{2}_{1},true),\;(y^{2}_{2},true)\}\big\}\\
R(s_1) &= &&\big\{(true),(false)\big\}\\
R(s_2) &= &&\big\{(false,true),(true,false)\big\}\\
\\
\end{array}$\\
We now impose the requirement \eqref{yInR}, that is 
$$(y^1 \in R^1(s_1)) \cup (y^2 \in R^2(s_2))$$
by adding it to $\mathcal G$. This requirement has earlier been defined as:
$$\mathcal G \gets \mathcal G \cup \left(\bigvee_{b \in R(s_1)} y^{1}_{1} = b_1\right) \cup \left(\bigvee_{b \in R(s_2)} y^{2}_{1} = b_1 \land y^{2}_{2} = b_2\right)$$
which corresponds to the requirement:
$$\phi(y^{1}_{1}) \in \{(true),(false)\} \textrm{ and } \phi(y^{2}_{1},y^{2}_{2}) \in \{(false,true),(true,false)\}$$ respectively for any $\phi \in sol(\mathcal D)$. The latter constraint removes the assignments:
$$\{(y^{1}_{1},false),(y^{2}_{1},true),(y^{2}_{2},true)\} \textrm{ and }
\{(y^{1}_{1},true),\;(y^{2}_{1},true),\;(y^{2}_{2},true)\}$$ from
$sol(\mathcal D)$. All constraints implied by the MDFAs are now
contained in $\mathcal D$

We now have $\vxi = \{w \in \Sigma^* \mid a^i(\dh_i (w)) \in \vyi\}$.
From this we get \\
$
\begin{array}{lll}
\vx 1 &= &\{w \in \Sigma^* \mid a(\dh_1 (w)) \in \vy 1\}\\
& = &\{w \in \Sigma^* \mid a(\dh_1 (w)) \in \{(true)\} \} \\
&=& L(\textrm{``a''})\\
\\
\textrm{and}\\
\\
\vx 2 &= &\{w \in \Sigma^* \mid a(\dh_2 (w)) \in \vy 2\}\\
& = &\{w \in \Sigma^* \mid a(\dh_2 (w)) \in  \{(false,true)\} \\
&=& L(\textrm{``abd*''})\\\\
\end{array}
$
\end{example}

\section{The Algorithms}\label{algorithms}
In this section we will present the three algorithms
\textsc{Build}$(\mathcal C)$, \textsc{Append}$(x_i,w)$ and
\textsc{ValidDomain}$(x_i)$. The first algorithm \textsc{Build}
constructs a data structure that is used by \textsc{Append} and
\proc{ValidDomain}. In all algorithms we assume that $\mathcal D, M_1,
\ldots, M_n, R^1,\ldots,R^n, a^1,\ldots,a^n,\Sigma$ and $\rho$ are
global variables. We assume that $\vyi$ is available. We
further assume that initially $\rho \gets
\{(x_1,\epsilon),\ldots,(x_n,\epsilon)\}, \mathcal G_2 \gets
\emptyset$ and $k_1,\ldots,k_n = 0$

\begin{codebox}
\Procname{\proc{Build}$(\mathcal C)$}
\li $\mathcal G_1 \gets \mathcal{F}$
\li \For $i \gets 1$ \To $n$
\li    \Do \For each $j$th match expression on the variable $x_i$ occuring in $\mathcal G_1$ as $match(x_i,\alpha^i_j)$
\li         \Do replace $match(x_i,\alpha^i_j) \textrm{ in } \mathcal G_1 \textrm{ by a new variable } y^{i}_{j}$
\li             $k_i \gets k_i +1$
\li         Build a DFA, $M'_{i,j}$ on $L(\alpha_{ij})$
\End
\li    $y^i = (y^{i}_{1}, \ldots, y^{i}_{k_i})$
\li    $M_i \gets \proc{ConstructMDFA}(M'_{i,1},\ldots, M'_{i,k_i})$ 
\li    $R^i \gets \proc{ComputeReachableAcceptanceStates}(M_i)$
\End
\li $\mathcal Y = \{y^{1}_{1},\ldots,y^{1}_{k_1},y^{2}_{1}, \dots y^{2}_{k_2}\, \ldots\ldots ,y^{n}_{1},y^{n}_{k_n}\}$
\li \For $i \gets 1$ \To $n$
\li    \Do  $\mathcal G_2 \gets \mathcal G_2 \cup \big(y_i \in R(s_i)\big)$. 
\End
\li $\mathcal D = (\mathcal Y,\mathcal G_1 \cup \mathcal G_2)$ 
\li \If $\vy 1 = \emptyset$
\li    \Then \Error ``No feasible solutions''
\End
\li \For $i \gets 1$ \To $n$
\li    \Do \For each $q_i \in Q_i$
\li       \Do \If $\{a^i(q_i)\} \cap \vyi = \emptyset$
\li              \Then $a^i(q_i) = \emptyset$
\End
\li           $R^i(q_i) \gets R^i(q_i) \cap \vyi$
\End
\li    \proc{Minimimize} $M_i$
\End
\end{codebox}

Line 1-10 constructs the first half of $\mathcal G$ based on $\mathcal
F$.  Line 11-12 constructs the second half of $\mathcal G$ and Line 13
defines $\mathcal D$. 14-15 check for feasible solution to $\mathcal
C$ the reason for using $\vyi$ instead of $sol(\mathcal D)$ is that
we have not required that $sol(\mathcal D)$ is available to us. Line 16-21 
tries to reduce the size of the data structure by removing the 
acceptance values from $a$ and $R$ that cannot lead to a valid solution.
Note that Line 18-19 might set $a(q) = \emptyset$, which is not valid 
according to the definition of an MDFA. However we use the value in the
pseudocode to indicate that this acceptance value never can be part of
a solution to $\mathcal D$.

\begin{codebox}
\Procname{$\proc{ValidDomain}(x_i)$}
\li $A_i \gets \emptyset$
\li \For each $q_i \in Q_i$ 
\li   \Do \If $a^i(q_i) \in \vyi$
\li       \Then $A_i \gets A_i \cup \{q_i\}$
\End\End
\li $\alpha \gets$ the regular expression corresponding to the DFA $(Q_i,\Sigma,\delta_i,s_i,A_i)$
\li \Return $\alpha$
\end{codebox}

This algorithm construct a DFA on the MDFA $M_i$ accepting 
$\vx i = \{w \in \Sigma^* \mid a(\dh_i(w)) \in \vyi \}$
and returns the regular expression corresponding 
to the constructed DFA. Of course we might consider other
ways to indicate the valid domains than by returning a regular
expression. This will be discussed in Section \ref{Extensions} 

\begin{codebox}
\Procname{\proc{Append}$(x_i,w)$}
\li $\mathcal G' \gets \mathcal G \cup \big(y^i \in R^i(\dhi(s_i,w))\big)$
\li \If $\mathcal G' \models \bot$
\li \Then \Error ``invalid append''
\End
\li $\rho(x_i) \gets \rho(x_i) w$
\li $s_i \gets \dh(s_i,w)$
\li $\mathcal G \gets \mathcal G'$
\end{codebox}

We append the letter $w$ to $\rho(x_i)$, and add a constraint to
$\mathcal G$ in order to remove the assignments on $\mathcal Y$
that are no longer possible to attain by any $\rho$. 

\section{Implementation}
In the algorithms we have supposed that we have a data structure on $\mathcal
D$ that supports two operations:
\begin{myenumerate}
\item Adding constraints to $\mathcal G$.
\item Computing $\vyi$ for every $1 \le i \le m$.

\end{myenumerate}
This could be done by filtering on $\mathcal G$ using one of the many
filtering approaches (see e.g. \cite{dechter}).  However in the
setting of interactive configuration, were values are assigned one by
one and valid domains and very fast valid domains computations has to
be available, encoding the constraints by a BDD seems to be the
obvious choice. We also choose to represent $R(q_i)$ as a BDD encoding
of he constraint $y_i \in R^i(q_i)$. Hence setting $\mathcal G \gets \mathcal G \cup
\big(y_i \in R(q_i)\big)$ can be done by setting
$\textrm{BDD}(\mathcal G) \gets \textrm{BDD}(\mathcal G) \land
\textrm{BDD}(y_i \in R(q_i))$, where BDD$(\mathcal H)$ is the
BDD-representation of the conjunction of the set of boolean formulas
in $\mathcal H$.

The algorithms used to minimize MDFAs in \proc{Build} is a direct
generalization of the one presented in \cite{AhoHopcroftUllman74}.
It runs in $|Q| \log |Q|$ when $Q$ are the states in the non-minimal MDFA.

The algorithm that transforms a DFA into a regular expression can be found
in \cite{hopcroft-ullman}. It runs in $O(|\delta|
\cdot |\alpha|)$ where $|\delta|$ is the number of transitions in the
DFA and $|\alpha|$ is the number of characters in the resulting
regular expression

\section{Extensions}\label{Extensions}
\subsection{Encompassing previous BDDs in the current context}
Since $\mathcal D$ is encoded as a BDD we can easily provide support
for boolean and integer variables allowing the same operations as
usual in on-line configuration. For instance we would be able to 
accept constraints as $x_2 \ne 7 \lor x_1 \land match(x_2,
"7*") \land match(x_3,"abc*")$ on the variables $x_1, x_2, x_3$.
Currently we cannot model equality of two string but it could easily
be added.

One might also choose to encode the integer as a
string in some cases. For instance a regular expression can be used to
determine whether a integer of infinite length is a factor of 2 or a
factor of 3.

\subsection{k-shortest path}
If we are to present the valid domain of a variable to the user, i.e.
to help the user in completing a string, a regular expression might
not be very intuitive -- especially if the concept of regular expressions 
is unknown for the user. Hence one might consider other strategies.

One idea would be only to output the shortest text-completion. This can
be done in $|Q| \log |Q| + |delta|$ using Dijkstras algorithm, where $|Q|$ and $|\delta|$ is the
number of states and transitions in the MDFA respectively.
We can also find the $k$ shortest paths
in $O(|\delta| + |Q| log |Q| + k)$ time \cite{KloopyPath} and find the
$k$ shortest simple paths in $O(k|Q|(|\delta|+|Q| log |Q|))$
\cite{KsimplePath}.

If more than one acceptance value is valid one might consider 
to output the shortest path to each of the valid acceptance 
values one at a time.

\subsection{Completing a string}\label{completingString}
We might want to support two kinds of updates:
\begin{itemize}
\item Appending a letter $w$ to a string $x_i \in \mathcal{X}$ as
earlier described
\item Completing a  string $x_i \in \mathcal{X}$
\end{itemize}

\noindent To complete a variable $x_i$ is in some way to state that no
more letters will be appended to $\rho(x_i)$. This could in the
example of input field validation be stated by the user in hitting the
return key or leaving a text field. We support this second update as
the action of appending a special letter \textsc{eol} $ \in \Sigma$ to
$\rho(x_i)$, and disallowing appending letters to $\rho(x_i)$ if the last
letter of $\rho(x_i)$ is \textsc{eol}.

\subsection{Making savings by a simple heuristic}
It might be considered to make a simple reduction. Rewritten
expressions like $match(x,\alpha) \lor match(x,\beta)$ to
$match(x,\alpha \cup \beta)$ and similarly $match(x,\alpha) \land
match(x,\beta)$ to $match(x,\alpha \cap \beta)$. These rewritings
may leads a large reduction in space as the DFA will not need to worry
about 2 cases instead of 4. 

\subsection{Supporting initial domain of $\mathcal X$}
In this paper we have assumed that the initial domain of any $x \in
\mathcal X$ is $\Sigma^*$. In practice we might want to constrain the
initial domain by a regular expression. For instance we might chose to
constrain the zip code to only contain digits from the very start by
adding
$match(\textsf{zip},\textrm{``}(0|1|2|3|4|5|6|7|8|9)*\textrm{''})$ to
$\mathcal G$ as an initial constraint.

\section{Future Work}
An obvious extension would be to explore whether it is possible to
achieve the same functionality with languages that are more expressive
than the regular languages. For instance we might investigate if we
can handle context-free languages \cite{hopcroft-ullman}.

Another thought that might be pursued is whether the input language
used to declare the constraints of $\mathcal F$ is appropriate for
declaring the constraints of $\mathcal F$. Formally it is perfect as
every regular language can be expressed as a regular expression.
However the length and complexity of these expressions may make it
cumbersome to express even simple constraints. Consider for instance
the constraint that $x$ is in the regular language of natural numbers
divisible by $3$. This regular language can be modeled by a DFA with 3
states and nine transitions. In our current inpu-language this will
have to be expressed as $f =
match(x,``([0369]*|([147]|([258][0369]*[258]))[0369]*([258]|([147][0369]*[147]))|([258]|([147][0369]*[147]))[0369]*([147]|([258][0369]*[258]))
)*$''$)$.  This suggest that we might consider some other ways to
model the DFA constraints than the $match$-expression. The ad hoc
solution to the problem stated above could be to allow expressions in
the input-language on the form ``$x $ modulo $ y = z$'' where $x, y, z
\in \mathbb Z$.  But we can easily construct similar problems that
will cause other problems. Hence a challenge is to consider how the
input language can be made in a way so that it is easy to express
problem the numerous problems that have nice DFAs but are horrible to
express as regular-expressions.

Another problem is how to make the user who in most cases will have
little or no acquaintance with regular expression make constraints
that can be enforced by the data structure.

\section{Acknowledgement}
We would like to thank Rasmus Pagh for useful discussions during the
making of this paper and Peter Tiedemann for thorough and
insight full readings of the paper, that resulted in many important
suggestions and corrections.

\newpage
\appendix
\begin{figure}[h]
\begin{center}
\input{verybigdfa.tex}\includegraphics[width=15.0cm]{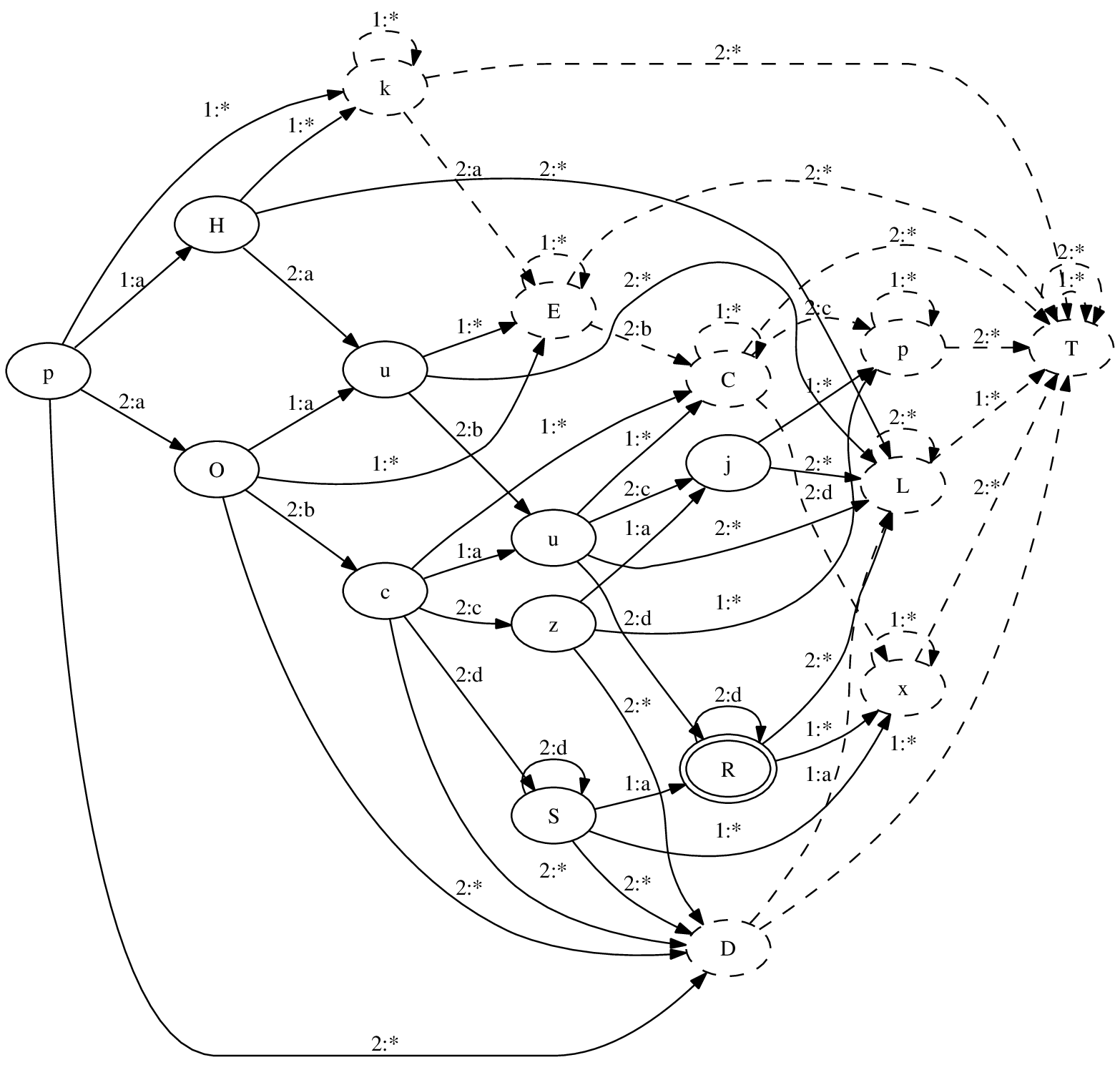}
\end{center}
\caption{A valid domains DFA built on the formulas $f_1 =
  match(x_2,$``abc''$) \lor match(x_1,$``a''$), f_2 =
  match(x_2,$``abd$*$''$)$. Transitions 1:* an 2:* means transitions
  on all other letter that cannot follow any transition on the first
  or second variable respectively. Dashed states are states where no
  accepting state is reachable. If the DFA is minimized they will all
  be contracted to the same state}.\label{veryBigDfa}
\end{figure}

\bibliography{regularBDD}

\end{document}